\def\year{2022}\relax
\documentclass[letterpaper]{article} % DO NOT CHANGE THIS
\usepackage{aaai22}  % DO NOT CHANGE THIS
\usepackage{times}  % DO NOT CHANGE THIS
\usepackage{helvet}  % DO NOT CHANGE THIS
\usepackage{courier}  % DO NOT CHANGE THIS
\usepackage[hyphens]{url}  % DO NOT CHANGE THIS
\usepackage{graphicx} % DO NOT CHANGE THIS
\usepackage{natbib}  % DO NOT CHANGE THIS AND DO NOT ADD ANY OPTIONS TO IT
\usepackage{caption} % DO NOT CHANGE THIS AND DO NOT ADD ANY OPTIONS TO IT
\DeclareCaptionStyle{ruled}{labelfont=normalfont,labelsep=colon,strut=off} % DO NOT CHANGE THIS
\usepackage{amssymb}
\usepackage{booktabs}
\usepackage{fancyhdr}
\usepackage{multirow}
\usepackage{algorithm}
\usepackage{algpseudocode}
\usepackage{amsmath}
\usepackage{graphics}
\usepackage{epsfig}
\usepackage{subfigure}

\usepackage{newfloat}
\usepackage{listings}
\floatstyle{ruled}
\newfloat{listing}{tb}{lst}{}
\floatname{listing}{Listing}
\title{Plug-Tagger: A Pluggable Sequence Labeling Framework Using Language Models}

\author{
    Xin Zhou\textsuperscript{\rm 1}\equalcontrib, Ruotian Ma\textsuperscript{\rm 1}\equalcontrib, Tao Gui\textsuperscript{\rm 2}\thanks{Corresponding authors.}, Yiding Tan\textsuperscript{\rm 1}, Qi Zhang\textsuperscript{\rm 1}\footnotemark[2], Xuanjing Huang\textsuperscript{\rm 1}

}

\affiliations{
    \textsuperscript{\rm 1}School of Computer Science, Fudan University, Shanghai, China \\
    \textsuperscript{\rm 2}Institute of Modern Languages and Linguistics, Fudan University, Shanghai, China \\
    \{xzhou20, rtma19, tgui, yidingtan20, qz, xjhuang\}@fudan.edu.cn
}
\usepackage{bibentry}
\begin{document}

\maketitle

\begin{abstract}Plug-and-play functionality allows deep learning models to adapt well to different tasks without requiring any parameters modified. 
Recently, prefix-tuning was shown to be a plug-and-play method on various text generation tasks by simply inserting corresponding continuous vectors into the inputs. 
However, sequence labeling tasks invalidate existing plug-and-play methods since different label sets demand changes to the architecture of the model classifier. In this work, we propose the use of label word prediction instead of classification to totally reuse the architecture of pre-trained models for sequence labeling tasks. Specifically, for each task, a label word set is first constructed by selecting a high-frequency word for each class respectively, and then, task-specific vectors are inserted into the inputs and optimized to manipulate the model predictions towards the corresponding label words. 
As a result, by simply switching the plugin vectors on the input, a frozen pre-trained language model is allowed to perform different tasks.
Experimental results on three sequence labeling tasks show that the performance of the proposed method can achieve comparable performance with standard fine-tuning with only 0.1\% task-specific parameters. 
In addition, our method is up to 70 times faster than non-plug-and-play methods while switching different tasks under the resource-constrained scenario. 
\end{abstract}

\section{Introduction}
\begin{figure}
    \centering
    \includegraphics[height=5.0cm, width=8.5cm]{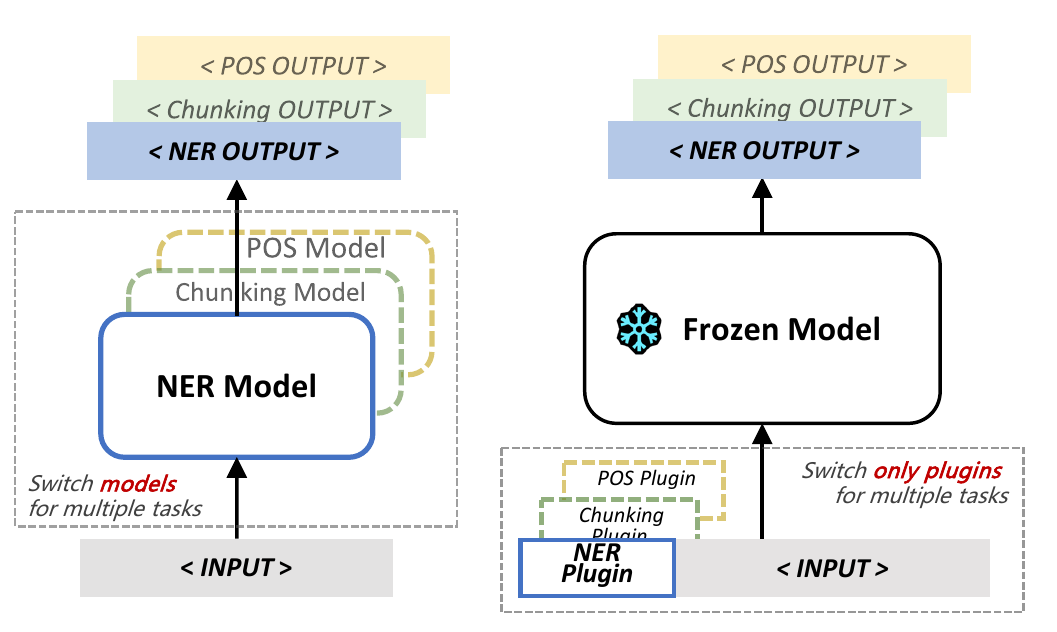}
    \caption{Comparison of fine-tuning (left) and plug-and-play method (right). When performing different tasks, the fine-tuning method needs to switch the model, while the plug-and-play method only needs to switch the plugin on input.}
    \label{fig:compare}
\end{figure}

% Fine-tuning is the primary paradigm of using pre-trained language models (PLMs) and performs well in many downstream NLP tasks \cite{radford2019language, devlin-etal-2019-bert}. However, fine-tuning requires retraining the large-scale parameters of PLM for each task, which can be expensive on a prohibitively large model like GPT3 \cite{brown2020language} (175B parameters). In real-world scenarios, even storing retrained PLMs like BERT (110M parameters) for each task can be tough, as shown in Figure \ref{fig:compare}(a). Therefore, it is necessary to find a better way to transfer the ability of PLMs to perform different tasks with minimal cost. OLD
Fine-tuning is a primary paradigm for transferring  pre-trained language models (PLMs) to many downstream natural language processing (NLP) tasks \cite{radford2019language, devlin-etal-2019-bert}. However, standard fine-tuning requires retraining the large-scale parameters of a PLM for each task. In real-world scenarios where tasks switch very frequently, retraining and redeploying \cite{8416871} the PLM can be prohibitively expensive. In addition, even storing PLMs like BERT (110M parameters) for each task can be tough. Therefore, it is necessary to find a better way to transfer PLMs towards different downstream tasks with minimal cost. 
In the ideal case, no modification of PLM is required, and performing different tasks does not require redeployment, but is done by switching lightweight modules. This approach is known as the plug-and-play method as shown in Figure \ref{fig:compare}.
% which can be prohibitively expensive on a very large model like GPT3 \cite{brown2020language} (175B parameters). 
% 在真实场景中存在着大量的新任务，为每个任务重新训练并重新部署PLM代价将会非常高昂，另一方面，
% In real-world scenarios, 
% In the ideal case, no parameter modification is required, different tasks require lightweight modules, known as plug-and-play. As shown in Figure \ref{fig:compare}.
% full plug-and-play behavior is achieved, and . 

% Adapter-tuning \cite{JMLR:v21:20-074,houlsby2019parameterefficient} and Prefix-tuning  \cite{li-liang-2021-prefix} are currently the most popular methods for alleviating these problems. These approaches focus on keeping most parameters of the PLMs fixed and optimizing a few parameters for each task. By inserting a task-specific adapter layer between layers of the PLM, adapter-tuning only needs to store one shared PLM and multiple task-specific layers to perform different tasks, rather than store full copies of PLM for each task. Prefix-tuning is the method to further alleviate the problems on generation tasks. It does not modify the model structure and instead prepends a sequence of continuous task-specific vectors to the input. These vectors work as a prompt for the model, allowing it to manipulate the generated language. Thus, tasks performed by the PLM can be switched by simply switching the prefix vector on the input. Prefix-tuning obtained good performance in different generation tasks and achieved plug-and-play. OLD

Adapter-tuning \cite{JMLR:v21:20-074,houlsby2019parameterefficient} and prefix-tuning \cite{li-liang-2021-prefix} are popular methods for achieving pluggability. These approaches focus on keeping parameters of the PLMs fixed and optimizing only a few parameters for each task. By inserting a task-specific adapter layer between layers of the PLM, adapter-tuning merely needs to store the parameters of one shared PLM and multiple task-specific layers to perform different tasks, rather than storing a full copy of the PLM for each task. 
Prefix-tuning does not modify the model architecture and instead prepends a sequence of continuous task-specific vectors to the inputs. These vectors work as a prompt for the model, manipulating it to generate outputs required by different tasks. Thus, generation tasks performed by the PLM can be switched by simply switching the prefix vectors at the inputs.

However, adapter-tuning adds additional layers to the PLM, which modifies the architecture, the inconvenience of redeploying the model when switching to new tasks remains unsolved.
Prefix-tuning is only plug-and-play on text generation because the language model head can be shared between different generation tasks, but different label sets make it impossible for classification tasks to share a classifier. Therefore, prefix-tuning cannot be directly transferred to classification tasks such as sequence labeling. 

To overcome this limitation, in this paper, we propose plug-tagger, a plug-and-play framework for various sequence labeling tasks. The proposed method allows a frozen PLM to perform different tasks without modifying the model but instead by switching plugin vectors on the input. The change to the label set leads to an unavoidable modification of the model classifier, but we overcome this problem by reformulating sequence labeling to the task of predicting the label words. We take high-frequency words predicted by PLM to serve as the label words for corresponding labels. For example, the entities of person names such as ``\emph{Xavier}" can be recognized by predicting a common name (label word) such as ``\emph{John}".
Benefiting from the reuse of the entire architecture of PLMs, our method can directly utilize plugin vectors to switch tasks without making any modifications to the model, as shown on the right side of Figure \ref{fig:compare}. Our method is also useful in industrial applications, where the PLM typically requires hardware acceleration for each new task deployment \cite{8416871}. A plug-and-play approach, with no modifications to the model architecture, allows us to accelerate multiple tasks with only one accelerated PLM.

The main contributions of this paper can be summarized
as follows: 

\begin{itemize}
\setlength{\itemindent}{1.1em}
    \item[$\bullet$] 
    We proposed a pluggable component on PLMs for sequence labeling, 
    and the proposed method can accomplish new tasks by modifying the input rather than modifying the model.
    \item[$\bullet$] 
    We propose a new paradigm for sequence labeling tasks. It transforms the classification task into a label word prediction task by reusing the language model head of PLM.
    \item[$\bullet$]
    Experiments on a variety of sequence labeling tasks demonstrate the effectiveness of our approach. Besides, in experiments with limited computational resources, our method is up to 70 times faster than other methods.
\end{itemize}

\section{Approach}
 \begin{figure*}
    \centering
    % 13.6*20
    \includegraphics[height=8.16cm, width=12cm]{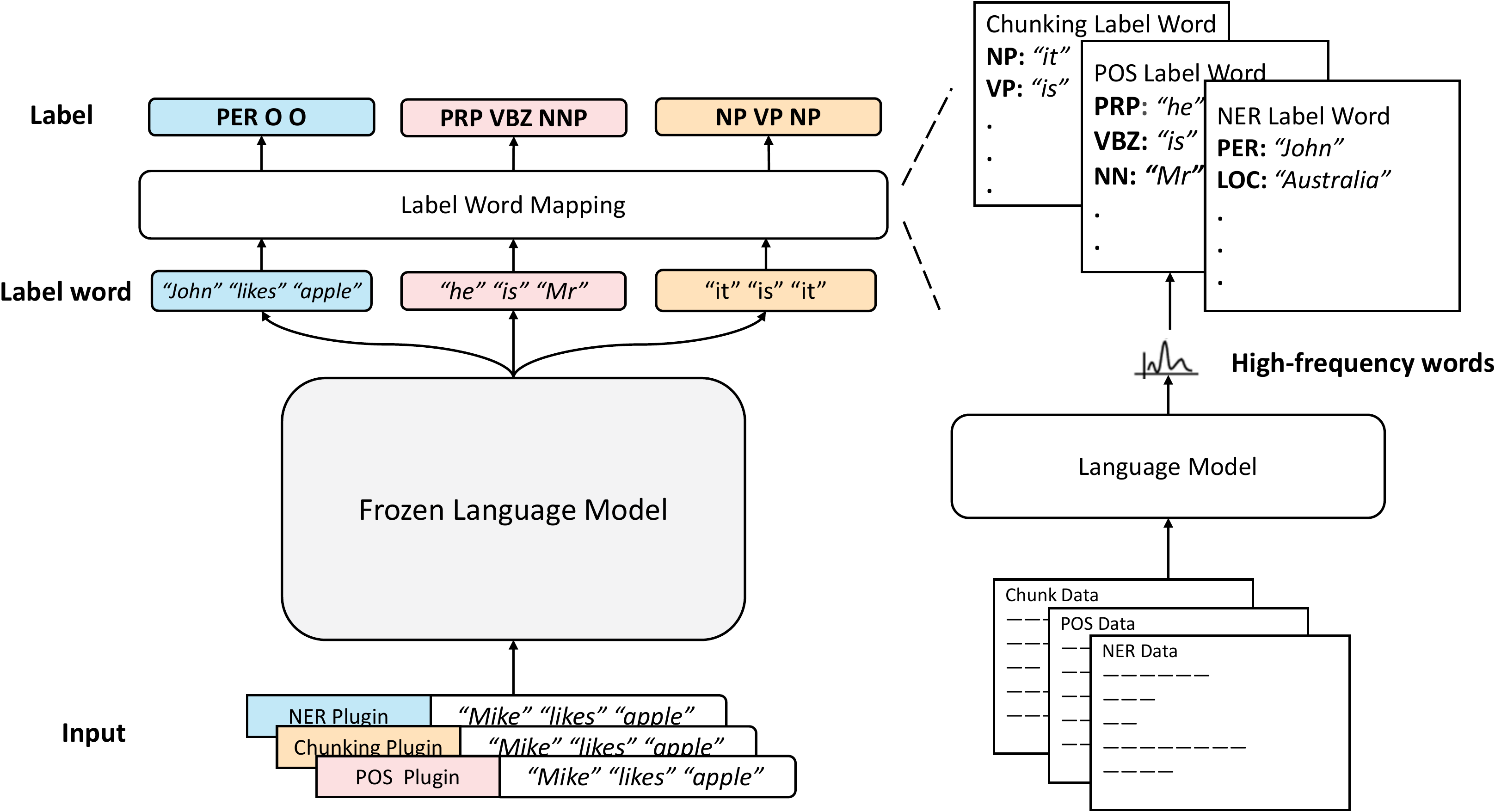}
    \caption{An overview of Plug-Tagger. 
    % The left side shows how we achieve plug-and-play. 
    Influenced by the task-specific plugin vector on the input, the frozen language model predicts the label word for each word in sentence. The actual label is obtained by label word mapping.
    % By inserting a task-specific vector into the input, the frozen language model is controlled to predict pre-found label words for each word in the sentence. The actual label is restored through label word mapping. 
   The right side shows how we get the label word: the language model traverses a large amount of data. The word that best represents a particular label is selected as a label word according to its frequency predicted by language model.
    }
    \label{fig:model}
\end{figure*}
In this work, we are looking for a plug-and-play method for sequence labeling, which aims to make a frozen model perform different sequence labeling tasks by switching task-specific lightweight modules on the input.
In this section, we first introduce the key challenges of achieving pluggability on sequence labeling. Next, we propose an overview of our approach, and finally, we detail the two primary components of our approach.
\subsection{Problem Statement}
Given a sequence of words $\bold{X} = [x_1,...,x_n]$, the goal of sequence labeling is to predict the sequence of gold label $\bold{Y} = [y_1,y_2,...,y_n]$ with equal length. 
% $\hat{Y} = F(X;\Theta)$
The predictions of a sequence labeling system can be expressed as $\bold{\hat{Y}} = F(\bold{X};\Theta)$ where $\Theta$ is the parameters of the system. 
The traditional system based on fine-tuning and classifier can be decomposed into the following equations:
% We can use the following equations to obtained the system's prediction:
\begin{equation}
\begin{split}
    & \bold{H} = Encoder_{\phi_1}(\bold{X}) \\
    & \bold{\hat{Y}} = argmax(Softmax(\bold{HW}+\bold{b})) \\
\end{split}
\end{equation}
where $\bold{W}\in\mathbb{R}^{h\times l}$, $l$ is the size of label set, $h$ is the dimension of hidden state. $Encoder_{\phi_1}$ is a PLM without language model head. Parameters of system $\Theta$ is decomposed into $\phi_1$, $\bold{W}$ and $\bold{b}$. 

The requirements of the plug-and-play method are lightweight and no modification of the model.
There are two challenges to achieve this goal.
The first challenge arises from the large-scale $\phi_1$, standard fine-tuning requires an entire new model for every task. That is, each task requires a large number of task-specific parameters when using PLM. The second one is that the dimension of $\bold{W}$ cannot be frozen due to different label sets. For example, NER's label consists of entity types such as person, location and organization, but POS's label consists of part of speech types like adjective, noun and adverb. It's challenging to map so many different task-specific labels onto the same label space. This prevents the model's classifier layer from remaining frozen.

\subsection{Model Overview}
    The architecture of the proposed model is shown in Figure \ref{fig:model}. 
    We switch the lightweight plugin vectors on the input rather than reloading large-scale parameters of PLM to perform different tasks. The label word mechanism replaces the task-specific classifier to avoid the modification of the model architecture. Under the influence of the plugin vector, the model predicts the corresponding label words of input, and the actual labels can be obtained by label word mapping. Take NER as an example, we feed the input ``\emph{Olivia likes apple}" with the plugin of NER into the frozen language model, the output of language model will be ``\emph{John likes apple}". ``\emph{John}" is the label word of PER (person) in NER, after label word mapping, we get the NER label of the input: ``PER O O".
    
    % For task $T=[t_1,...,t_m]$, $[\Theta_{t_1},...,\Theta_{t_m}]$ is the parameters of all task, the traditional system can be written as  $F(X;\Theta_T)$, 
    
    We define the label word map as $M$, $\Theta_{lm}$ is the parameters of the frozen language model, $\Theta_P$ is plugin vector.
    The label words predicted by plug-tagger can be describe as $\tilde{Y} = F(\{X, \Theta_P\};\Theta_{lm})$ where $\{X, \Theta_P\}$ is the inputs, and we use label word mapping to get the real labels $\hat{Y} = M(\tilde{Y})$.
    The following two sections detail essential parts of the plug-tagger: plugin vector and label word mechanism.
 
\subsection{Plugin Vector}
    To solve the problem of storing and reloading large-scale parameters for various tasks, we plugin vector to control the model behavior without modifying the architecture and parameters of the model. 
    The plugin vector $\Theta_P$ consists of a few continuous vectors and is combined with input.
    In the following two subsections, we show two ways to combine the plugin vector with the input.
\subsubsection{Plugin in embedding}
    The input of PLM is the text $\bold{X}$ processed by embedding, which can be described as $\bold{X}' = [\bold{x_1}',...,\bold{x_n}']$ where $\bold{x_i}'=Emb(x_i)$. 
    The plugin vectors can be described as $\Theta_P=[\theta_1,...\theta_{l_p}]$ where $\Theta_P \in \mathbb{R}^l_p \times\ h$ , $l_p$ is the length of the plugin vectors and $h$ is the dimension of embedding.
    The plugin vectors are inserted into the input $X$ directly, the new input can be described as follow:
    \begin{equation}
        \bold{X}' = [\Theta_P;\bold{x}_1',...,\bold{x}_n'],
    \end{equation}
    % The plugin vectors $\Theta_P$ to the input after embedding layer to obtain $\bold{X}'=[\Theta_P; \bold{x_1}', ..., \bold{x_n}']$.  
    where $\Theta_P \in \mathbb{R}^{l_p \times h}$, [;] means concatenation in the first dimension. PLMs take $\bold{X}'$ as input, the information in $\Theta_P$ flows through each layer and ultimately affects the predictions. 
    % During training,  $\Theta_P$ is the only parameters that are optimized.
\subsubsection{Plugin in layer}
    The plugin vectors inserted to embedding are not expressive enough. To extend the influence of plugin vectors, we insert them into input for each layer of the model. 
    Given a PLM with $l$ transformer layers, the input of $j^{th}$ layer can be described as $\bold{X}^{(j)} = [\bold{x}_1^{(j)},...,\bold{x}_n^{(j)}]$ where $\bold{X}^{(j)} \in \mathbb{R}^{n \times d}$, $d$ is the dimension of hidden state and and $n$ is the length of inputs.
    Transformer layer are structured around the use of query-key-value (QKV) attention, which is calculated as:
    \begin{equation}
    \begin{split}
    Att(\bold{X}) = Soft&max(\frac{Q(\bold{X})K(\bold{X})^T}{\sqrt{d_k}})V(\bold{X}) \\
    Q(\bold{X}) &= \bold{W}_q^{(j)}\bold{X} \\
    K(\bold{X}) &= \bold{W}_k^{(j)}\bold{X} \\
    V(\bold{X}) &= \bold{W}_v^{(j)}\bold{X},
    \end{split} 
    \end{equation}
    % $Attention(\bold{X})=softmax(\frac{\bold{W}_q\bold{X}(\bold{W}_k\bold{X})^T}{\sqrt{d_k}}) \bold{W}_v\bold{X}$, 
    where $\bold{W_k}^{(j)}, \bold{W_v}^{(j)}, \bold{W_q}^{(j)} \in \mathbb{R}^{d \times d}$ and $d_k$ is the  number of multi head. In order to avoid adding additional layers of PLM, we combine the plugin vector with $K$ and $V$, which can be describe as:
    \begin{small}
    \begin{equation}
    \begin{split}
    &\bold{X}' = \{\bold{X}, \theta_k^{(j)}, \theta_v^{(j)}\} \\
    % \theta_k^{(j)} = MLP(\theta_k^{(j)}) \\
    % \theta_k^{(j)}
    Att(\bold{X}') = Softm&ax(\frac{Q(\bold{X})[\theta_k^{(j)};K(\bold{X})]^T}{\sqrt{d_k}})[\theta_v^{(j)};V(\bold{X})],
    \end{split} 
    \end{equation}
    \end{small}
where $\theta_k^{(j)}, \theta_v^{(j)} \in \mathbb{R}^{l_p \times d}$. Plugin vectors on all layers can be represented as $\Theta_P = \{(\theta_k^{(1)},\theta_v^{(1)})...,(\theta_k^{(l)},\theta_v^{(l)})\}$. We extend the influence of plugin vectors to every layer without introducing new parameters of PLM.

\subsection{Label Word Mechanism} \label{section:label word}
    To alleviate problems caused by the task-specific classifier, we propose label word mechanism, which reformulates the sequence labeling to the label word prediction. Label word is the most common word for a label.  For example, the uncommon name ``\emph{Olivia}" can be substituted with the more common name ``\emph{John}".
    % We believe that replacing uncommon words with common words in the same category not affect the person to label the sequence. 
    % has little impact on sequence labeling.
    % If each label can be represented by a word in vocabulary, we can reuse the language model head to perform different tasks, instead of replacing the classifier layer.
    % Label word mechanism consists of two steps:

    \subsubsection{Label Word Selection}
 
    Algorithm \ref{alg:label word} represents the entire label word selection processing. 
    For each label, a dictionary $freq$ was built to counts its candidate label words and corresponding frequency.  We the traverse training set, for each word in the sentence, we use the language model to get top-k high-probability candidate words and update dictionary $freq_c$ corresponding to the word's label $c$. After traversing the training set, we filter some words that are not suitable for label words such as the words that frequently occur in all $freq$. Under the condition that the label word of each label is not the same, the remaining word with the highest frequency in $freq_{c}$ was selected as label word of label $c$. 

    In particular, for tasks that need to use the BIO schema, two special treatments are needed: 1) We don't count label word for label O. It's hard to pick a representative word for the others category. In the training and inference phase, the word with label O predict itself. 2) We look for label words respectively for B and I of the same category, because the internal order of the phrases brings more information, and distinguishing BI is beneficial for tasks that require boundary information.
    In the experiment section, we will discuss the influence of the label word mechanism on the performance in detail.
    \begin{algorithm}[t]
    \caption{Label Word Selection} %算法的名字
    \label{alg:label word}
    \hspace*{0.02in} {\bf Input:} %算法的输入， \hspace*{0.02in}用来控制位置，同时利用 \\ 进行换行
    Train set $D=\{X_i,Y_i\}_{i=1}^N$; Label set $L=\{c_i\}_{i=1}^l$; Vocabulary $V=\{w_i\}_{i=1}^v$; Pre-trained language model $LM$; Maximum candidates of label word $k$  \\ 
    \hspace*{0.02in} {\bf Output:} %算法的结果输出
    LabelMap $M$ 
    \begin{algorithmic}[1]
    \State Initialize label map $M=\varnothing$;
    \For{$c \in L$}
        \State Initialize $freq_c = \{w_i:0\}_{i=1}^v$;
        \State Add label word pair $\{c:None\}$ to $M$;
    \EndFor
    \For{($X=\{x_i\}_{i=1}^n,Y=\{y_i\}_{i=1}^n$) $ \in D$} % For 语句，需要和EndFor对应
        \State Select top-$k$ candidate words $\{\tilde{y}_i\}_{i=1}^n$ where $ \tilde{y}_i \in \mathbb{R}^k$ based on predictions of language model $LM(X)$;
        \For{$i \in [1...n]$} 
        \State Update the frequency of label $c=y_i$;
        \State $freq_{c}[w] \leftarrow freq_{c}[w]+1$ for $w \in \tilde{y}_i$;
        \EndFor
    \EndFor
    \For{$c \in L$}
        \State Filter out irrelevant words in $freq_c$;
        \While{$M[c]$ is $None$}
            \State Select the word $w$ in $freq_c$ with the highest frequency;
            \If{$w$ not used by $M$}
                \State $M[c] \leftarrow w $;
            \Else
                \State Remove $w$ from $freq_c$;
            \EndIf
        \EndWhile
    \EndFor
    
    % \While{condition} % While语句，需要和EndWhile对应
    % \EndWhile
 \State \Return M
    \end{algorithmic}
    \end{algorithm}
    \subsubsection{Training objective}
    We reformulate sequence labeling to a special language modeling task. After selecting label words, we get the label map $M$ to map label set to words in vocabulary. For sequence $X$, the gold label Y is reintroduced to $\widetilde{Y} = [\widetilde{y}_1,...,\widetilde{y}_n]$ where $\widetilde{y_i} = M(y_i)$. 
    The sentence-level loss can be described as follows:
    \begin{equation}
    \begin{split}
    Loss = -\sum\limits_{i=1}^{N}{log(P(\widetilde{Y_i}|X_i))}, \\
    \end{split} 
    \end{equation}
  where N is the number of sentences. When combined with the plugin vectors, parameters of plugin vectors $\Theta_P$ and the embeddings of label words are the only trainable parameters. 
   During the inference phase, we take the prediction result of the first subword of each word and find its corresponding label according to the label map.

\section{Experiments}
In this section, we present the experimental results to show the transferability and efficiency of plug-tagger.
We conducted experiments on three sequence labeling tasks to verify whether plug-tagger could adapt to different tasks. We also simulate scenarios that require redeployment to verify whether plug-tagger could ease the inconvenience caused by redeployment.
% We conducted experiments on three sequence labeling tasks: 
\subsection{The transferability of plug-tagger}
\subsubsection{Dataset}
To verify the transferability of our methods, we conduct experiments on NER, POS and chunking. 
CoNLL 2003 shared task (CoNLL2003) \cite{sang2003introduction} is the standard benchmark dataset which was collected from Reuters Corpus.
This dataset consists of annotations for NER, POS and chunking, so we conduct three tasks on it. In addition, we selected another representative dataset for each task, including ACE 2005 (ACE05) \footnote{https://catalog.ldc.upenn.edu/LDC2006T06} for NER, Wall Street Journal (WSJ) \cite{marcus-etal-1993-building} for POS and CoNLL 2000 \cite{tjong-kim-sang-buchholz-2000-introduction} for chunking. We use the BIO2 tagging scheme for all the tasks and follow the standard dataset preprocessing and split. The CoNLL 2000 does not have an officially divided validation set, we use the test set as the validation set. The statistics of the datasets are summarized in Table \ref{tab:datasets}.
\begin{table}[htbp]
    \centering
    \renewcommand\arraystretch{1.2}
    % \resizebox{\textwidth}{!}
    % {
    \setlength{\tabcolsep}{2.5mm}
    \scriptsize
    \begin{tabular}{cccccc}
    \toprule
    Task & Dataset & \#Train & \#Dev & \#Test & Class  \\
    \midrule
    \multirow{2}{*}{NER} & CoNLL 2003 & 204, 567 & 51, 578 & 46, 666 & 8\\
     & ACE 2005 & 144, 405 & 35, 461 &30, 508 & 14 \\
     \hline
    \multirow{2}{*}{POS}  & CoNLL 2003 & 204, 567 & 51, 578 & 46, 666 & 45 \\
     & WSJ & 912, 344 & 131, 768 &129, 654 & 46 \\
     \hline
    \multirow{2}{*}{Chunking}  & CoNLL 2003& 204, 567 & 51, 578 & 46, 666 & 20 \\
     & CoNLL 2000 & 211, 727 & - &47,377 & 22 \\
    \bottomrule
    \end{tabular}
    % }
    \caption{Statistics of the datasets on NER, POS and chunking. \# means number of tokens in dataset.}
    \label{tab:datasets}
\end{table}
\subsubsection{Baseline Methods}
\begin{table*}[]
    \centering
    \renewcommand\arraystretch{1.5}
    % \resizebox{\textwidth}{!}{
     \setlength{\tabcolsep}{1.5mm}{
     \begin{tabular}{cc|cc|cc|cc}
    \toprule
      \multirow{2}{*}{Params} & \multirow{2}{*}{Method}  & \multicolumn{2}{c|}{NER (F1)}& \multicolumn{2}{c|}{Chunking (F1)} & \multicolumn{2}{c}{POS (Acc.)} \\
        & &  CoNLL2003 & ACE2005 & CoNLL2003 & CoNLL2000&CoNLL2003 &WSJ \\
    \midrule
    100\% & FINE-TUNING & $91.45$ & $89.02$ & $91.41$ & $97.05$ & $95.64$  & $97.69$ \\
    \hline
    
    \multirow{2}{*}{0.01\% - 0.1\%}
    & FT-Sub & $84.27_{-7.18}$ & $77.37_{-11.65}$ & $79.94_{-11.47}$ & $83.13_{-13.92}$ & $88.97_{-6.67}$ &$94.94_{-2.75}$ \\
    % & Adapters & - & - & - & - & - & - \\
    & Plug-Tagger & $\bold{87.68_{-3.77}}$ & $\bold{79.98_{-9.04}}$ & $\bold{84.99_{-6.42}}$ & $\bold{92.02_{-5.03}}$ & $\bold{93.28_{-2.36}}$ & $\bold{96.73}_{-0.96}$ \\
    \hline 
    \multirow{2}{*}{0.1\%-1\%} 
    &  Adapters & $88.89_{-2.56}$ & $\bold{88.03_{-0.99}}$ & $88.52_{-2.89}$ & $94.63_{-2.42}$ & $93.51_{-2.13}$ & $97.51_{-0.18}$ \\
    & Plug-Tagger & $\bold{91.58_{+0.13}}$ & $87.68_{-1.34}$ & $\bold{90.50_{-0.91}}$ &$\bold{96.48_{-0.57}}$ & $\bold{95.01_{-0.63}}$ & $\bold{97.60_{-0.09}}$ \\
  
    \bottomrule
    \end{tabular}
      }
    \caption{Experimental results for all datasets on the three tasks. The Params represent the percentage of task-specific parameters required by the method.  The number after each result means the difference between this result and FINE-TUNING. Bold terms mean the best result in the same scale of parameters.}
    \label{tab:main_result}
\end{table*}
We compare plug-tagger with three baselines:
\begin{itemize}
\setlength{\itemindent}{1.1em}
    \item[$\bullet$] \textbf{FINE-TUNING} \cite{devlin-etal-2019-bert} optimizes the all parameters of PLM with a task-specific classifier. Standard fine-tuning can show the normal performance of a pre-trained model.
    % FT-sub 我们用A来展示不添加额外结构和参数的PLM
    \item[$\bullet$] \textbf{FT-Sub} only optimizes sub-layer of PLM, which is the most straightforward lightweight method. We use FT-Sub show the performance of lightweight PLM without additional layers and parameters.
    to limit task-specific parameters to the same scale as plug-tagger.
    \item[$\bullet$] \textbf{Adapters} \cite{houlsby2019parameterefficient} is a well-known parameter-efficient method which optimizes the parameters of additional layers inserted in PLM. We control the scale of the Adapters' parameters by varying the number of adapter layers and the dimension of the middle layer.
\end{itemize}
\subsubsection{Experiment Details}
We employ Roberta-Base \cite{liu2019roberta} as our base model. Plug-tagger and all baselines and are based on Roberta-base.
The parameters and architecture are reloaded directly from HuggingFace\footnote{https://huggingface.co/}. 
For adapter-tuning, we use the Adapter-Hub\footnote{https://github.com/Adapter-Hub/adapter-transformers} that combines adapter-tuning and Transformers released by \cite{pfeiffer2020AdapterHub}. 
We keep the default optimizer and scheduler settings, AdamW, and linear scheduler are used for all datasets. 
The maximum length of the tokenizer is 128. The default epoch is set to 10, and the batch size is 16.
We find that a large learning rate often results in better performance when optimizing vectors on the input. So the default learning rate was set to 1e-3 for plug-tagger and prefix-classifier and 1e-5 for others. The hyperparameters used by each method on all data sets are given in the appendix.

For plugin vector, we additionally tune the length of plugin vectors. All experiments were conducted under the same random seed. The results are based on the hyperparameters, which are selected based on the performance on the validation set. 
% Hyperparameter details are in the appendix.
% 

As for label words, We use the language model to obtain the high-frequency words of a label on the training set, and according to the combination of high-frequency words of different labels, we can select a variety of label words. The results reported in this paper are based on the set of label words with the best performance on the validation set. 

To compare different methods at the same parameter scale. We limit the number of layers and size of FT-Sub and Adapters, the details are presented in the appendix.

\subsubsection{Main Results}
% 在不同参数规模下baseline的性能。在只用0.1%的参数时，Plug-tagger展现出了和ft相近的性能, and outperform adapters in a large margin
Tabel \ref{tab:main_result} presents the results of all methods on the test set under different parameter scales. With only 0.1\% task-specific parameters, plug-tagger almost outperform all other lightweight methods at the same parameter scale and achieve a comparable performance with FINE-TUNING. Under the setting of 0.01\% parameters, our method performs worse than than FINE-TUNING, but still better than other methods. Next, we analyze the experimental results in detail according to the task.

\textbf{NER} is the most difficult task in our experiments. We find that our method outperform FINE-TUNING on CoNLL2003 and has the largest performance drop of all data set in ACE2005, we believe this is due to the more complex scenario of ACE2005, and less data in ACE2005 makes it more difficult to find the appropriate label words. \textbf{Chunking}'s label has the widest range of label word candidates, which brings challenges for finding suitable label words. But we see only a small drop in chunking's two datasets comparing to FINE-TUNING, suggesting that the plug-tagger can handle this situation. \textbf{POS} has the largest number of labels.  The good performance in CoNLL2003 and WSJ proves that plug-tagger can adapt to many label words. 
\subsection{The efficiency of plug-tagger}
    \begin{table}[htbp]
    \centering
    \renewcommand\arraystretch{1.2}
    \setlength{\tabcolsep}{0.7mm}
    % \scriptsize
    \small
    \begin{tabular}{ccccc}
    \toprule
    Method   & Performance & Parameter & Architecture & NoRedeploy  \\
    \midrule
    Fine-tuning & $\checkmark$  & $\times$ & $\times$& $\times$   \\
    Classifier    &  $\times$  &  $\times$  &  $\times$& $\times$   \\
    Adapter  & $\checkmark$ & $\checkmark$ &  $\times$& $\times$  \\
    Plug-Tagger  & $\checkmark$ & $\checkmark$ &  $\checkmark$&  $\checkmark$ \\
    \bottomrule
    \end{tabular}
    \caption{Comparison of the degree of modification of the model by different methods when switching tasks. Performance means the method can achieve a comparable performance. Parameter\&Architecture means that the model does not need to modify the model parameter\&Architecture. NoRedeploy means the method can perform different tasks without redeployment.}
    \label{tab:speed}
\end{table}

In many edge devices, computational resources are minimal\cite{liu2021hardware}. It is not easy to run two large parameter models simultaneously, which leads to the necessity of releasing the resources of the old model to reload the new model when switching tasks, a process we call redeployment. Redeployment takes much time, and since it is unknown what tasks users need to complete in real scenarios, the frequency of switching tasks can be very high, which makes the disadvantages of redeployment even more apparent. As shown in Table 3, since plug-tagger switches tasks via the plugin vector on the input instead of changing the model's architecture and parameters, there is no need to redeploy the model when switching tasks.
To demonstrate the impact of redeployment, we designed an experiment that approximates a real scenario. For the real scenario where the task type is unknown, we sampled data from NER, POS, and chunking, mixing and disrupting the sampled results.  To simulate the case of unknown task types, we set the batch size to 1 to make the model predict only the current sample. To simulate the resource-constrained case, if the current task is different from the previous one, the model releases its parameters of the previous task and then reloads the new task-specific parameters. We count the time it takes to complete the tasks for all sampled samples in order. The data for each task is sampled from CoNLL2003.
We call the method that requires the model to be redeployed Model-Switch, and the method of simply loading the plugin vector without modifying the model is called Plugin-Switch.
All experiments are conducted in the same NVIDIA GeForce RTX 3090.
\begin{figure}
    \centering
    \includegraphics[height=7cm, width=9cm]{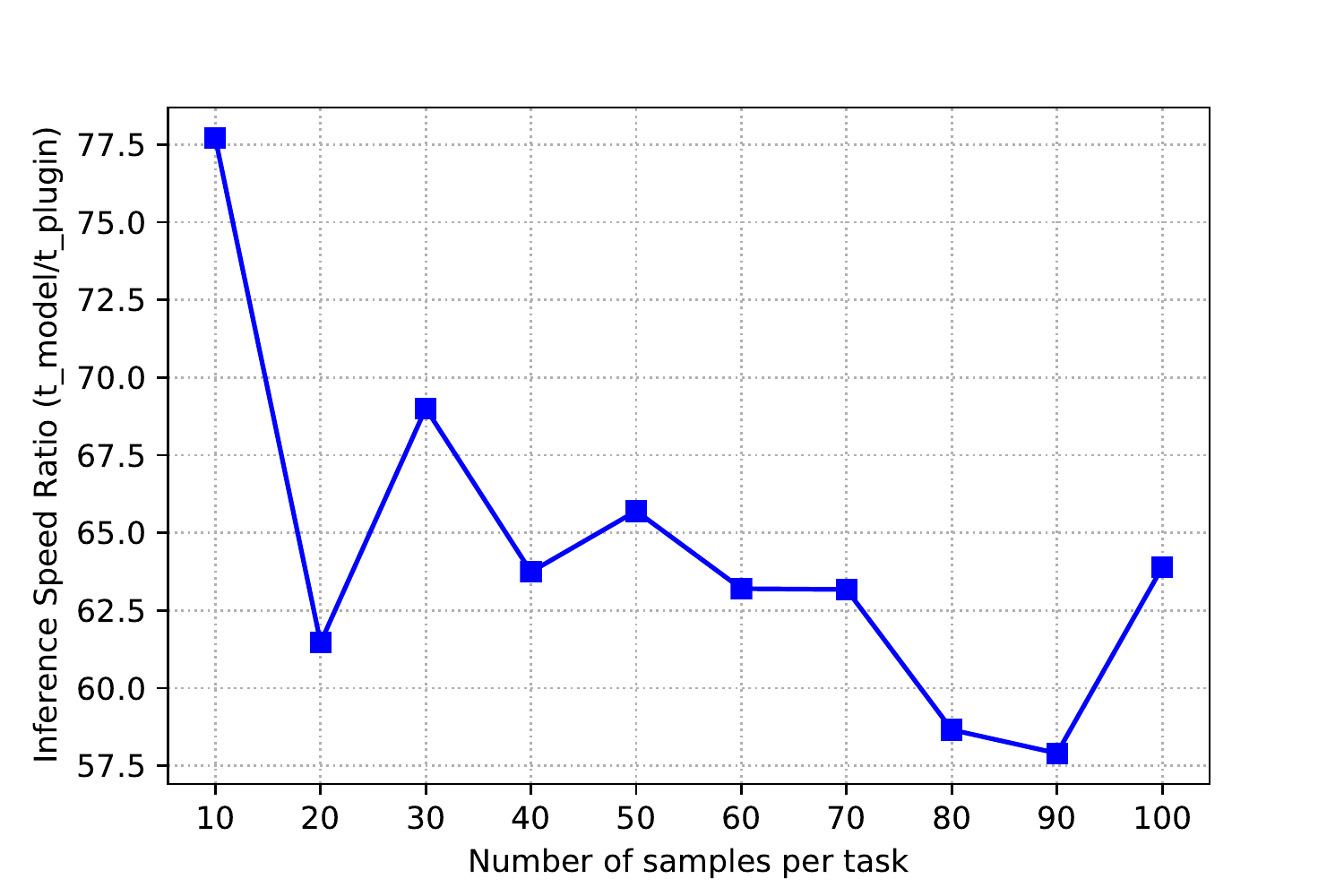} % 
    \caption{
    Inference speed ratio (inference time of plugin-switch method divide inference time of model-switch method) calculated on three sequence labeling task. }
    \label{fig:speed}
\end{figure}

We find that model-switch method fails miserably in this setting. It takes 367 seconds to process 100 samples per task. To reflect that gap, we calculate how many times faster the plugin-switch method is than the Model-switch method, as shown in Figure \ref{fig:speed}. The plugin-switch method is always much stronger than the model-switch method as the sample size increases. As can be observed in the figure, the inference speed ratio of the Plugin-switch method drops as the number of samples increases,  We believe this is because increasing the sample size increases the probability of consecutive identical tasks, reducing the frequency of task switching. Despite this, our method is at least 50 times faster than the model-Switch method. This proves that the plug-and-play approach works better in resource-constrained scenarios than model-swtich methods.

\subsection{Impact of label word mechanism}
In this section, we explore the impact of the label word mechanism on downstream tasks, and discuss the reason of the results.
\subsubsection{Comparison of language model head and classifier}
  \begin{table}[htbp]
    \centering
    \renewcommand\arraystretch{1.5}
    % \setlength{\tabcolsep}{1.2mm}
    % \scriptsize
    \small
    \begin{tabular}{cccc}
    \toprule
    Task&Dataset&Plug-Classifier & Plug-Tagger \\
    \midrule
    \multirow{2}{*}{NER} &CoNLL2003& 89.52 & 91.58 \\
    &ACE2005&85.49 & 87.68 \\
    \hline
    
    \multirow{2}{*}{Chunking} &CoNLL2003& 88.8 & 90.5 \\
    &CoNLL2000&94.39 & 96.48 \\
    \hline
    
    \multirow{2}{*}{POS} & CoNLL2003& 92.45&95.01 \\
    &WSJ&96.05&97.6 \\
    \bottomrule
    \end{tabular}
    \caption{Performance of Plug-Tagger and Plug-Classifier under the same setting. Plug-Classifier is the combination of plug vectors and classifier. Plug-Tagger is the combination of plug vectors and the language model head.}
    \label{tab:classifier}
\end{table}
Plugin vectors can be adapted to the classifier directly, Exploring this method in the main experiment makes no sense because it is not a plug-and-play method. In this subsection, we discuss whether the label word mechanism has a negative effect on downstream tasks compared to the classifier. 
We write the combination of plugin vector and classifier as plug-classifier. 
For a fair comparison, 
we evaluate the performance of the plug-classifier on all datasets with the same hyperparameters as the plug-tagger.
% we use the same hyperparameters to compare the performance of plug-classifier and plug-tagger on all data sets. 
The experimental results are shown in Table \ref{tab:classifier}. We found that the label word mechanism is much better than the classifier on all datasets. 
We believe the performance benefits from 1) The task of predicting words is similar to the tasks of pre-training. 2) The label word mechanism uses more pre-training parameters because of the reusing of the language model head. These two reasons make the label word mechanism more suitable for prompt-based methods. Therefore, plug vector is better suited to be combined with the label word mechanism.

\subsubsection{Comparison of label word option}
% plugin vector也可以和classifer结合，但该方法并不具备可拔插的属性。

Recall in section Approach, we discuss the option of selecting label words. We compare two ways to select a label word for the BIO schema. One is to select a label word for B label, and I label respectively. When performing label Word mapping, label words can be matched directly, so this option is called ExactMatch. The other is that select a label word to represent both the B label and I label. During the inference phase, adjacent and identical label words are combined, the first label word is considered B, and the rest are considered I. We called this method GreedyMatch.

\begin{figure}
    \centering
    \includegraphics[height=5cm, width=8cm]{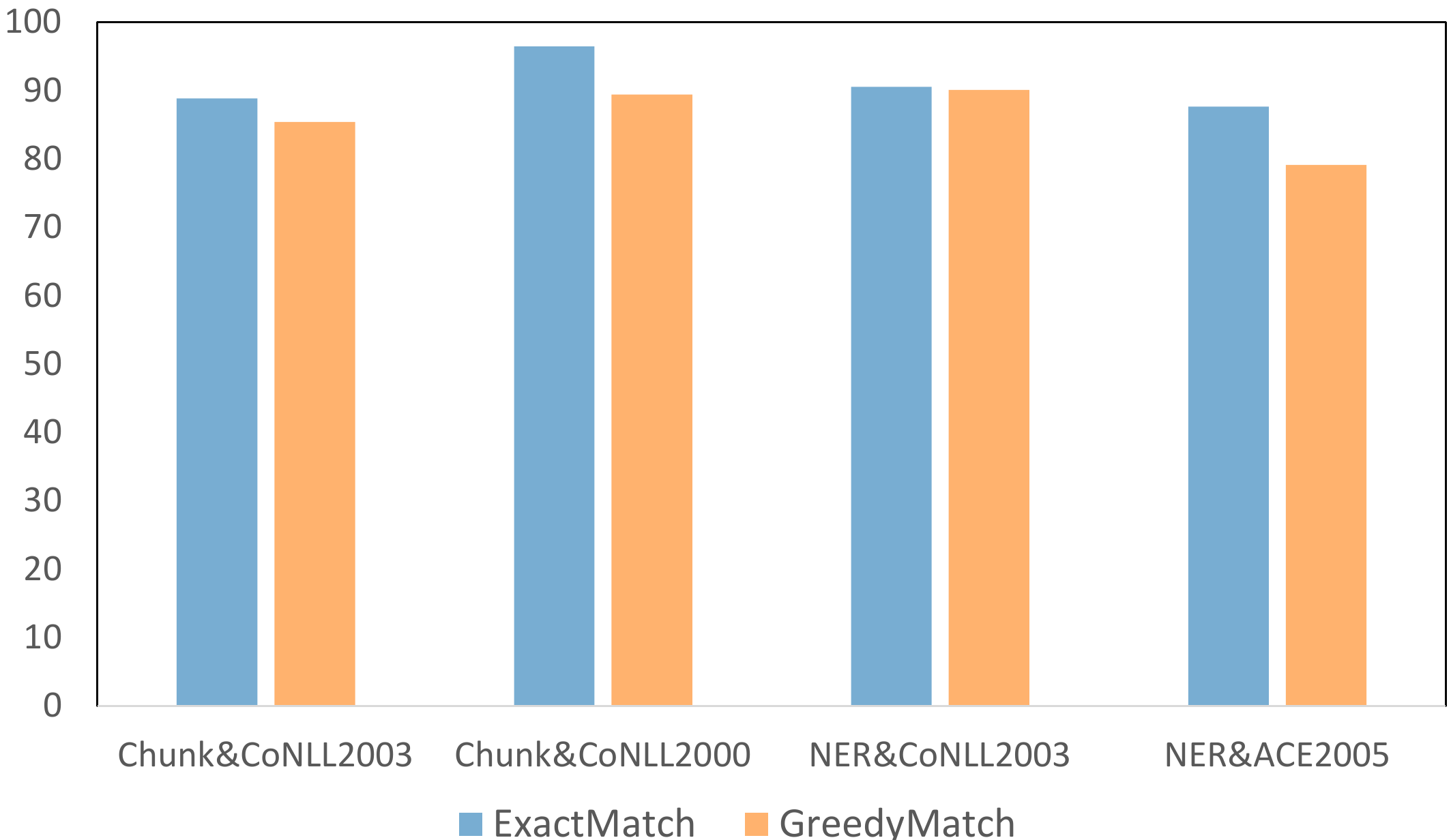}
    \caption{Performance comparison of two label word options in datasets based on BIO tagging schema.}
    \label{fig:label word}
\end{figure}

As shown in Figure \ref{fig:label word}, we find that GreedyMatch works better on the NER task of CoNLL2003 but otherwise performs worse than ExactMatch on all other datasets, especially the performance degradation is very obvious for each dataset of chunking. We found that there are adjacent words of the same label in the chunking task, but they represent two different phrases and therefore cannot be combined.
For NER, the same phenomenon occurs in ACE2005, but it is not found in CoNLL2003. This leads to inconsistencies in the trends of the two NER datasets. 
We suppose that the following two reasons cause GreedyMatch to improve NER's performance on CoNLL2003: 1) We find that there are no adjacent similar entities in CoNLL2003. 2) Some simple labels may be better represented by the same word. But GreedyMatch cannot be widely used because there are not many scenarios for which it is suitable. ExactMatch is the more practical option since it achieved better performance in more situation.

% \subsection{Impact of plugin vectors length}

\section{Related Work}
\subsection{Sequence labeling}

Sequence labeling, such as named entity recognition (NER), part-of-speech (POS) tagging and chunking, is one of the fundamental tasks of natural language processing \cite{ma2016end}. 
Early work for sequence labeling foused on feature engineering for
graphical models like conditional random fields (CRFs) and Hidden Markov Models(HMM) \cite{lafferty2001conditional, passos2014lexicon, cuong2014conditional}. 
Recently neural network models achieve competitive performances \cite{chiu2016named, dos2014learning, luo2020hierarchical}, and fine-tuning the pre-trained language models  \cite{devlin-etal-2019-bert, liu2019roberta, yang2020xlnet} have been shown to achieve state-of-art results on sequence labeling  \cite{bell-etal-2019-context}. The above approach basically treats sequence labeling as token-level classification and requires a task-specific classifier, works of  \cite{athiwaratkun-etal-2020-augmented, yan-etal-2021-unified-generative} convert sequence labeling into a generation task, avoiding classifier by using the sequence-to-sequence framework \cite{NIPS2014_a14ac55a, cho-etal-2014-learning, vaswani2017attention, lewis-etal-2020-bart}. 
% 序列标注模型需要
But this work still needs to modify the model architecture, the inability to use native PLM and the inefficiencies of sequence-to-sequence limit its transferability and efficiency, which is not suitable for achieving our goals.

% 介绍预训练和常用的预训练范式
\subsection{Pre-trained language model}
Self-supervised representation models \cite{radford2018improving, radford2019language, yang2019xlnet, peters2018deep, devlin-etal-2019-bert} have shown substantial advances in natural language understanding after being pre-trained on large-scaled text corpora. Given an NLP task, the mainstream paradigm to use PLM is finetuning, which stacks a linear layer on top of the pretrained language model and then updates all parameters of model. Recently, prompt-tuning was received much attention. To get the better use of knowledge in pre-trained models,  \cite{han2021ptr, schick2021exploiting,cui-etal-2021-template, wang2021entailment, chen2021knowprompt, liu2021gpt} reformulate the paradigm of downstream tasks into new task that similar to pre-training. The standard fine-tuning relies on classifiers, prompt-tuning lacks an efficient approach to for sequence labeling. None of these paradigm suits our needs.
% Our method does not add the task-specific classifier for to perform different task, but instead predicts the label words for all sequence labeling tasks.

\subsection{Lightweight deep learning}
Lightweight fine-tuning method aims to using small trainable parameters to leverage the ability of PLMs \cite{houlsby2019parameterefficient}.
Some studies argue that redundant parameters in the model should be deleted or masked \cite{zaken2021bitfit, sanh2020movement, zhao2020masking, frankle2019lottery}, while others argue that additional structures should be added to the model \cite{zhang2020sidetuning, houlsby2019parameterefficient, guo2021parameterefficient}. For example, adapter-tuning insert some additional layer between each layers of PLMs. Plug-and-play method controls model behavior without making any modifications to the model parameters \cite{madotto2020plugandplay}, which is also a lightweight mehtod. \cite{li-liang-2021-prefix} and \cite{hambardzumyan2021warp} is a plug-and-play and lightweight method. It inserts continuous vectors into the input to allow a fixed PLM perform different tasks. But these works can not adapted to sequence labeling directly. 

% 有些工作的做法时modify decoding procedure: 1)re-weighting the output distribution \cite{}

% \clearpage
% Lightweight fine- tuning freezes most of the pretrained parameters and modifies the pretrained model with small trainable modules. The key challenge is to identify high-performing architectures of the modules and the subset of pretrained parameters to tune. One line of research considers removing parameters: some model weights are ablated away by training a binary mask over model parameters (Zhao et al., 2020; Radiya-Dixit and Wang, 2020). Another line of research considers inserting parameters. For example, Zhang et al. (2020a) trains a “side” network that is fused with the pretrained model via summation; adapter-tuning inserts task-specific lay- ers (adapters) between each layer of the pretrained LM (Houlsby et al., 2019; Lin et al., 2020; Rebuffi et al., 2017; Pfeiffer et al., 2020). Compared to this line of work, which tunes around 3.6\% of the LM parameters, our method obtains a further 30x reduction in task-specific parameters, tuning only 0.1\%
% 介绍Adapter和prefix-tuning
% 

\section{Conclusion}
In this work, we propose plug-tagger, a plug-and-play framework for sequence labeling. The proposed framework can accomplish different tasks using vectors inserted into the input and a fixed PLM without modifying the model parameters and architecture. It achieves competitive performance on the sequence annotation tasks with only a small number of parameters and is 70 times faster than other methods in real-world scenarios requiring model redeployment.
% Use \bibliography{yourbibfile} instead or the References section will not appear in your paper
\bibliography{aaai22}
\end{document}